\documentclass[10pt,twocolumn,letterpaper]{article}

\usepackage{cvpr}
\usepackage{times}
\usepackage{graphicx}
\usepackage{amsmath}
\usepackage{amssymb}

\usepackage[nocompress]{cite}
\usepackage{subcaption}
\usepackage{adjustbox}
\usepackage{booktabs}
\usepackage{threeparttable}
\usepackage{multirow}
\usepackage{algorithm}
\usepackage{algorithmicx}
\usepackage{algpseudocode}
\usepackage{enumitem}

\usepackage{color}
\definecolor{white}{RGB}{200,200,200}
\definecolor{black}{RGB}{40,44,52}
\definecolor{darkred}{rgb}{0.8,0,0}
\definecolor{darkgreen}{rgb}{0,0.7,0}
\definecolor{darkblue}{rgb}{0,0.0,0.9}

\usepackage[pagebackref=true,breaklinks=true,letterpaper=true,colorlinks,bookmarks=false]{hyperref}
\hypersetup{
  linkcolor=darkred,
  citecolor=darkgreen,
  urlcolor=darkred,
  pdfinfo={
    Title={Adversarial Feature Alignment: Avoid Catastrophic Forgetting in Incremental Task Lifelong Learning},
    Author={Xin Yao, Tianchi Huang, Chenglei Wu, Rui-Xiao Zhang, Lifeng Sun},
  }
}

\cvprfinalcopy 

\def\cvprPaperID{1} 

\begin{document}

\title{Adversarial Feature Alignment: Avoid Catastrophic Forgetting\\ in Incremental Task Lifelong Learning\thanks{Published in \textit{Neural Computation} Volume 31, Issue 11 (\href{https://www.mitpressjournals.org/doi/full/10.1162/neco_a_01232}{Link}).}}

\author{Xin Yao, Tianchi Huang, Chenglei Wu, Rui-Xiao Zhang, Lifeng Sun\\
Department of Computer Science and Techonology \\
Tsinghua University, Beijing, China\\
{\tt\small \{yaox16,htc19,wucl18,zhangrx17\}@mails.tsinghua.edu.cn, sunlf@tsinghua.edu.cn}
}

\maketitle

\begin{abstract}
Human beings are able to master a variety of knowledge and skills with ongoing learning.
By contrast, dramatic performance degradation is observed when new tasks are added to an existing neural network model.
This phenomenon, termed as \emph{Catastrophic Forgetting}, is one of the major roadblocks that prevent deep neural networks from achieving human-level artificial intelligence.
Several research efforts, e.g. \emph{Lifelong} or \emph{Continual} learning algorithms, have been proposed to tackle this problem.
However, they either suffer from an accumulating drop in performance as the task sequence grows longer, or require to store an excessive amount of model parameters for historical memory, or cannot obtain competitive performance on the new tasks.
In this paper, we focus on the incremental multi-task image classification scenario.
Inspired by the learning process of human students, where they usually decompose complex tasks into easier goals, we propose an adversarial feature alignment method to avoid catastrophic forgetting.
In our design, both the low-level visual features and high-level semantic features serve as soft targets and guide the training process in multiple stages, which provide sufficient supervised information of the old tasks and help to reduce forgetting.
Due to the knowledge distillation and regularization phenomenons, the proposed method gains even better performance than finetuning on the new tasks, which makes it stand out from other methods.
Extensive experiments in several typical lifelong learning scenarios demonstrate that our method outperforms the state-of-the-art methods in both accuracies on new tasks and performance preservation on old tasks.
\end{abstract}

\section{Introduction}

Conventional deep neural network~(DNN) models customized for specific data usually fail to handle other tasks, even when they share a lot in common.
When facing new tasks, these complicated models have to be either retrained or finetuned.
A typical example of retraining the model for new tasks is joint training, where new layers or other components are added to the existing model every time a new task arrives and then the whole model is trained on all the datasets.
However, such a solution not only incurs substantial data storage costs but also requires training the model from scratch for every new task, which is inevitably time- and computation-consuming.
The latter method, i.e., finetuning, is more prevalent in practice since it just retrains the model with new task data and then enables the old model to produce new task results.
It works well on the latest task but suffers from a decline of performance on previous tasks, which is termed as \emph{Catastrophic Forgetting}~\cite{goodfellow2013empirical}.

\begin{figure}
  \centering
  \includegraphics[width=\linewidth]{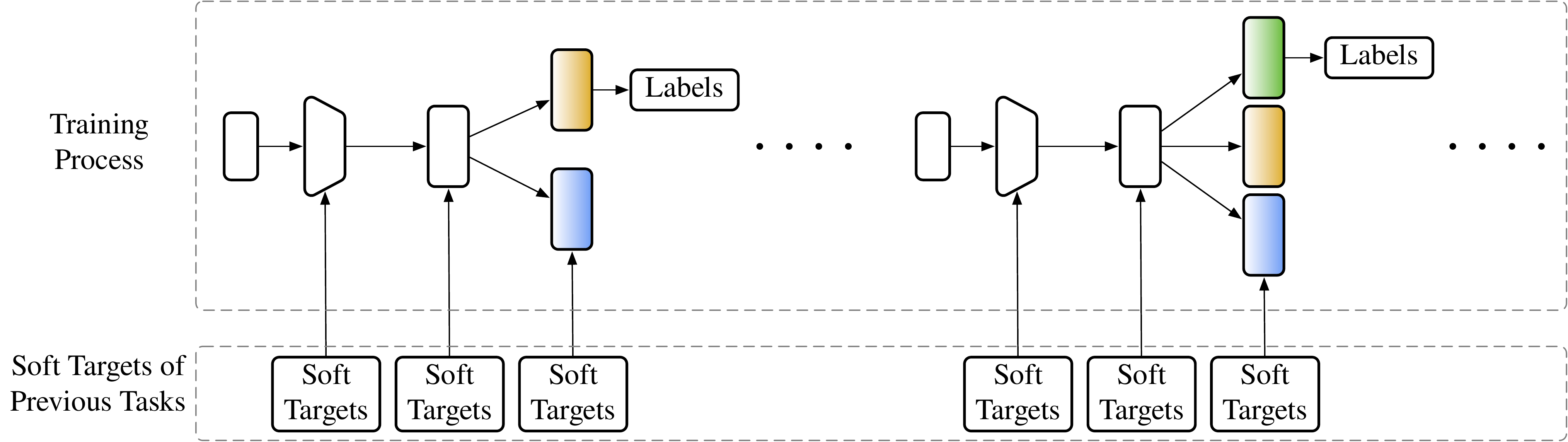}
  \caption{Lifelong learning with multilevel soft targets. They provide sufficient supervised information of the old tasks, and act as regularizers for the new tasks meanwhile.}
  \label{fig:multilevel}
\end{figure}

To address this problem, various lifelong learning algorithms \cite{rusu2016progressive,kirkpatrick2017overcoming,li2017learning} have been proposed during the past few years, in which they aim to preserve the performance on previous tasks while adapting to new data.
These methods try to alleviate the forgetting problem from different perspectives and can be categorized as follows, all of which bear some weakness (detailed information is provided in Section~\ref{sec:lls}):
architectural strategies require storing an increasing amount of model parameters while rehearsal ones need to store part of the training samples, both violating the motivation of lifelong learning to some extent;
among regularization strategies, parameter regularization strategies do not perform well when the tasks have different output domains or start from a small dataset, as revealed by our analysis and experiments;
activation regularization methods are reported to suffer a buildup drop in old tasks' performance as the task sequence grows longer~\cite{aljundi2017expert}.

In this work, we focus on the incremental multi-task image classification scenario.
Formally, we aim to retrain an existing model and enable it to perform well on all the tasks added in separately and sequentially, without access to the legacy data or storing an excessive amount of model parameters.
We find that current activation regularization strategies usually use the classification probabilities of previous models as soft targets and try to preserve old tasks' performances via knowledge distillation~\cite{hinton2015distilling}.
However, the classification probabilities cannot provide as strong supervised information as the hard targets (i.e., the labels) of the new tasks.
Thus the models are likely to deviate from the optimal points of previous tasks when adapting to the new data.
Meanwhile, we observe that in human life, when a student tries to learn something difficult, it is usually useful to decompose the complex task to several easier gradual goals.

Inspired by these insights, we propose a novel activation regularization method to alleviate the forgetting problem.
Since the \emph{activations} (denoting the outputs of neural network layers in this paper) of the old model have integrated the knowledge of previous tasks, they are able to guide the training process in multiple stages by serving as multilevel soft targets.
The typical training process with multilevel soft targets is illustrated in Figure \ref{fig:multilevel}.
Nevertheless, it is challenging to characterize the value of these intermediate activations.
For example, the convolutional features are in high dimensions and contain spatial structure information, while the fully-connected features are rich in abstract semantic information.
To tackle the problem, we novelly introduce a trainable discriminator to align the low-level visual features while applying Maximum Mean Discrepancy (MMD)~\cite{gretton2012optimal} to the high-level semantic features.
The overall algorithm is termed as Adversarial Feature Alignment~(AFA).
On the one hand, feature alignment provides sufficient supervised information of the old tasks during training and thus helps to alleviate the forgetting problem.
On the other hand, it distills the knowledge of the old models to the new one and acts as a regularizer for the new tasks, which enables the model to gain even better performance than finetuning on the new tasks.

To summarize, our contributions are as follows:
\begin{itemize}
\item We propose a novel activation regularization lifelong learning method to alleviate the forgetting problem via adversarial feature alignment, which not only preserves the performance on the old tasks but also achieves better performance on the new tasks.
\item We propose to align convolutional attention maps with a trainable discriminator and high-level semantic features with MMD. They guide the training process in multiple stages and helps to reduce forgetting, while guaranteeing better performance on the new tasks due to knowledge distillation and regularization.
\item Extensive experiments in several typical lifelong learning scenarios demonstrate that our method outperforms the state-of-the-art methods in both accuracies on the new tasks and performance preservation on old tasks.
\end{itemize}

\section{Related Work}

\begin{figure}
  \centering
  \includegraphics[width=0.98\linewidth]{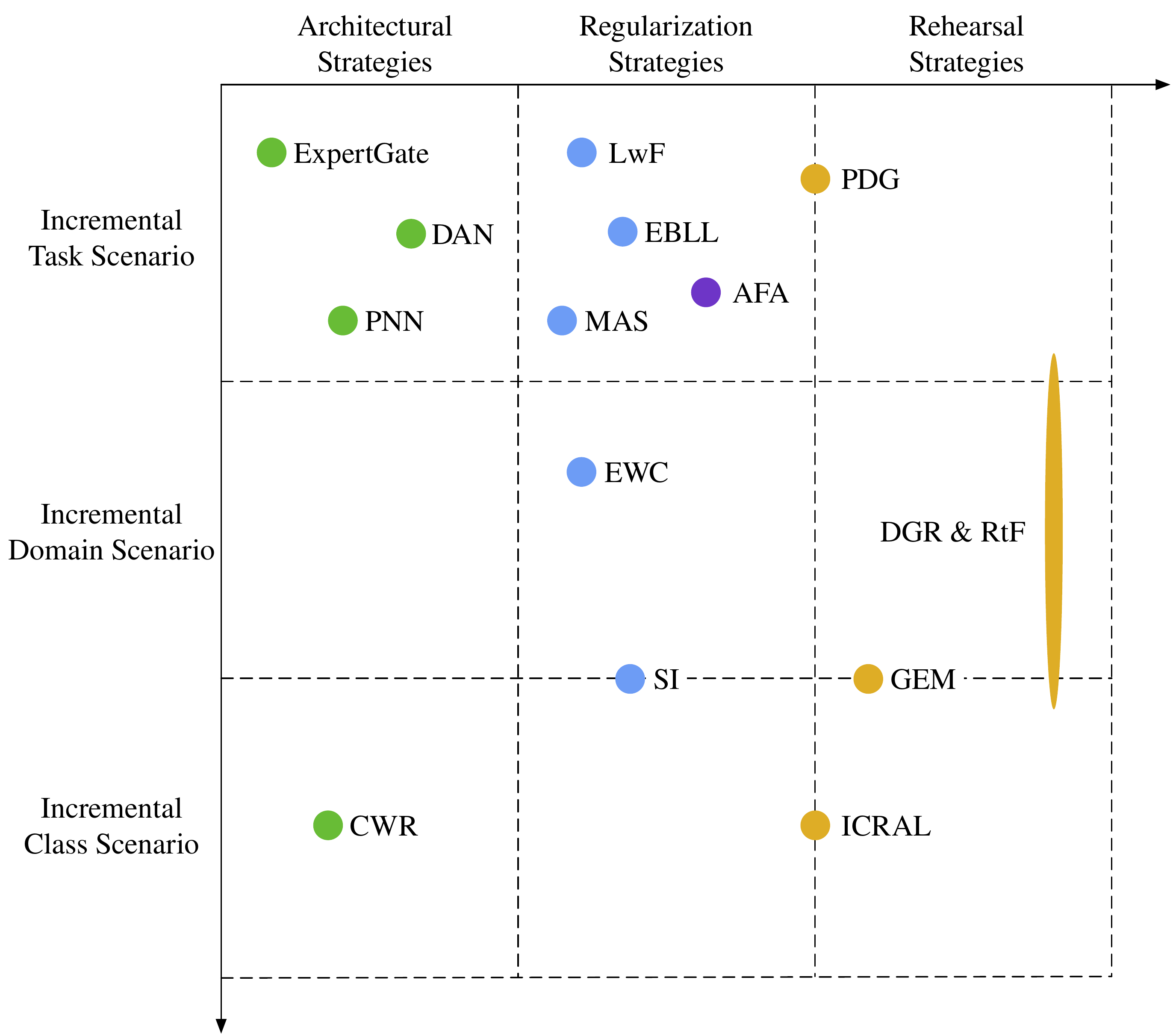}
  \caption{Strategies and scenarios of the representative lifelong learning methods: PNN~\cite{rusu2016progressive}, ExpertGate~\cite{aljundi2017expert}, DAN~\cite{rosenfeld2017incremental}, CWR~\cite{lomonaco2017core50}, GEM~\cite{lopez2017gradient}, ICARL~\cite{rebuffi2017icarl}, PDG~\cite{hou2018lifelong}, DGR~\cite{shin2017continual}, RtF~\cite{van2018generative}, LwF~\cite{li2017learning}, EBLL~\cite{rannen2017encoder}, EWC~\cite{kirkpatrick2017overcoming}, SI~\cite{zenke2017continual}, MAS~\cite{aljundi2018memory} and AFA, hereby proposed. (Better viewed in color)}
  \label{fig:strategies}
\end{figure}

Lifelong learning, sometimes called continual learning or incremental learning, has been attracting more research interests in recent years.
However, there is no standard terminology to identify various strategies or experimental protocols.
One method performs well in some experimental settings may totally fail in others~\cite{kemker2018measuring}.
For a fairer and more structured comparison, we combine the categorization of lifelong learning strategies in~\cite{maltoni2018continuous} and the distinct experimental scenarios in~\cite{van2018generative}, and then illustrate the representative lifelong learning methods as in Figure \ref{fig:strategies}.

\subsection{Lifelong Learning Strategies}
\label{sec:lls}

\subsubsection{Architectural strategies}
Architectural strategies train separated models for different tasks, and usually, a selector is introduced to determine which model to launch during inference.
Progressive Neural Network (PNN)~\cite{rusu2016progressive} is one of the first architectural strategies, in which a layer is connected to both the previous layer of the current model and the layers of old models, allowing information to flow horizontally.
Expert Gate~\cite{aljundi2017expert} decide which expert model to launch during inference and which training strategy to apply for the new incoming task with auto-encoders.
Other architectural strategies include Incremental Learning through Deep Adaptation (DAN)~\cite{rosenfeld2017incremental}, Copy Weight with Re-init (CWR)~\cite{lomonaco2017core50}, etc.

Architectural methods enjoy an advantage of preserving the performance on old tasks, as adding new tasks into the system will not harm the previously learned models.
However, an increasing amount of extra spaces are needed to store parameters for each task, which conflicts with the original intentions of lifelong learning.

\subsubsection{Rehearsal strategies}
Rehearsal strategies replay past information periodically while adapting to the new data to avoid forgetting.
Gradient Episodic Memory (GEM)~\cite{lopez2017gradient} uses a fixed memory to store a subset of old patterns.
Incremental Classifier and Representation Learning (ICARL)~\cite{rebuffi2017icarl} and Progressive Distillation and Retrospection (PDG)~\cite{hou2018lifelong} includes an external memory to store a subset of old task data.
These two methods also employ a distillation step and thus overlap with the regularization strategies.

Inspired by the recent success of Generative Adversarial Networks (GAN)~\cite{goodfellow2014generative}, Deep Generative Replay (DGR)~\cite{shin2017continual} and Replay through Feedback (RtF)~\cite{van2018generative} propose training generative models to generate samples w.r.t. previous task distributions.
However, training generative models for complex images is still an open problem itself.
Current generative replay strategies are limited to simple experimental settings, e.g., MNIST~\cite{lecun1998gradient} and its variations~\cite{goodfellow2013empirical}.

\subsubsection{Regularization strategies}
Regularization strategies extend the loss function with additional terms to preserve the performance on previous tasks.
It can be further divided into two categories: \emph{activation regularization} strategies and \emph{parameter regularization} ones.

\emph{Activation regularization strategies} are usually based on knowledge distillation and use additional loss terms between the activations of models to maintain the performance on old tasks.
Learning without Forgetting (LwF)~\cite{li2017learning} propose using outputs of the old models as soft targets of old tasks.
These soft targets are considered as a substitute for the data of previous tasks, which cannot be accessed in lifelong learning settings.
Encoder-based Lifelong Learning (EBLL)~\cite{rannen2017encoder} prevents the reconstructions of convolutional features from changing with auto-encoders, which has the effect of preserving the knowledge of the previous tasks.

\emph{Parameter regularization strategies} focus more on the model itself.
They try to figure out the importance of parameters for old tasks and apply a penalty to the change of essential parameters during training on new tasks.
Elastic Weights Consolidation (EWC)~\cite{kirkpatrick2017overcoming} estimates the importance of parameters by the diagonal of the Fisher information matrix and uses individual penalty for each previous task.
Synaptic Intelligence (SI)~\cite{zenke2017continual} estimates the importance weights in an online manner by the parameter specific contribution to the changes in the total loss.
Memory Aware Synaptic (MAS)~\cite{aljundi2018memory} is similar to SI but estimates the importance weights by the changes of the model outputs.
Parameter regularization strategies pay more attention to preserving the knowledge on old tasks but prevent the model from achieving competitive performance on the new tasks.

\subsection{Lifelong Learning Scenarios}

Benchmarking lifelong learning lacks a universal standard even if we focus on supervised image classification and leave out reinforcement learning, as various experimental protocols have been used in previous researches.
For a fairer comparison, we follow the distinct scenarios for lifelong learning proposed in~\cite{van2018generative} and summarize the experimental settings of the above strategies.

\subsubsection{Incremental Task Scenario}
In the incremental task scenario, the tasks are similar but have different output domains, e.g., all the tasks belong to image classification, but one cares about fine-grained flowers while another may focus on fine-grained birds.
Architectural strategies~\cite{rusu2016progressive,aljundi2017expert,rosenfeld2017incremental} can easily handle this scenario by training a specific model for each task, while some other methods~\cite{li2017learning,rannen2017encoder,hou2018lifelong, aljundi2018memory} usually introduce a \textit{multi-head} output layer w.r.t. each task in this scenario.

\subsubsection{Incremental Domain Scenario}
In the incremental domain scenario, the tasks usually have the same output domain but follow different data distributions.
Typical examples of such protocol include MNIST~\cite{lecun1998gradient} to SVHN~\cite{netzer2011reading}, both being digits classification but collected in different manners; and permuted MNIST~\cite{goodfellow2013empirical}, in which different permutations are applied to the pixels and each permutation corresponds to a unique task.
GEM~\cite{lopez2017gradient} and some parameter regularization strategies~\cite{kirkpatrick2017overcoming,zenke2017continual} mainly evaluate their algorithms in such a scenario.

\subsubsection{Incremental Class Scenario}
The last one is the incremental class scenario, where the model is required to learn to recognize new classes continually.
An example of such protocol is learning to classify MNIST (split MNIST) or CIFAR10~\cite{krizhevsky2009learning} (split CIFAR10) class by class.
These two datasets are widely adopted in CWR~\cite{lomonaco2017core50}, ICARL~\cite{rebuffi2017icarl}, GEM~\cite{lopez2017gradient}, SI~\cite{zenke2017continual}, etc.

Generative replay strategies~\cite{shin2017continual, van2018generative} conduct experiments in all the above scenarios.
However, due to the complexity of generative models, they have not shown the ability to apply to more complicated datasets other than MNIST.

In this paper, we do not consider lifelong learning methods that require storing model parameters (i.e., architectural strategies) or training samples (i.e., rehearsal strategies).
Concretely, we compare with the state-of-the-art regularization strategies (both activation and parameter based) in the incremental task scenario.

\section{Background}

\subsection{Problem Definition and Notations}
We first briefly introduce the lifelong learning setup and notations under the incremental task scenario, as used in \cite{li2017learning, aljundi2017expert, rannen2017encoder, hou2018lifelong, aljundi2018memory}.

Formally, let $X^{t} = \{x^t_i\}^{N^t}_i$ and $Y^{t} = \{y^t_i\}^{N^t}_i$ be the inputs and corresponding labels from task $t$ with $N_t$ denoting the number of examples.
Lifelong learning algorithms focus on how to transfer an existing model to $\{X^{\tau}, Y^{\tau}\}$ when task $\tau$ arrives without access to samples of the previous task, i.e., $\{X^t, Y^t\}^{\tau-1}_{t=1}$.
In other words, they try to train a model that performs well on a sequence of tasks incoming separately and sequentially with only the most recent data.
In this work, we focus on supervised image classification and the model is usually a Convolutional Neural Network (CNN).
Following the notations in~\cite{rannen2017encoder}, the model for task $t$ is denoted as $f_t$ and can be decomposed as $C_t \circ C \circ F$ where:
\begin{itemize}
\item $F$ is the feature extractor, e.g., the convolutional layers in CNNs.
\item $C$ is the shared classifier, e.g., the fully connected layers in CNNs except the last one.
\item $C_t$ is the task-specific classifier for task $t$, e.g., the last fully connected layer in CNNs.
\end{itemize}

\subsection{Formulation of Existing Methods}
\label{sec:existing}

If all the data of the previous tasks were available, we could train a model to handle multiple tasks with joint training by minimizing the following empirical risk:
\begin{equation}
\label{eq:joint1}
\sum^{\tau}_{t=1} \frac{1}{N_t} \sum^{N_t}_{i=1} \ell_{ce}(f_t(x^t_i), y^t_i) = \sum^{\tau}_{t=1} \mathbb{E}[\ell_{ce}(f_t(X^t), Y^t)]
\end{equation}
Considering the decomposition of $f_t$, Eq. (\ref{eq:joint1}) can be written as:
\begin{align}
\label{eq:joint2}
\mathcal{L}_{joint} &= \sum^{\tau}_{t=1} \mathbb{E}[\ell_{ce}(C_t \circ C \circ F(X^t), Y^t)] \\
&= \mathcal{L}_{cls} + \mathcal{L}_{old} \\
\label{eq:cls}
\mathcal{L}_{cls} &= \mathbb{E}[\ell_{ce}(C_\tau \circ C \circ F(X^\tau), Y^\tau)] \\
\mathcal{L}_{old} &= \sum^{\tau-1}_{t=1}\mathbb{E}[\ell_{ce}(C_t \circ C \circ F(X^t), Y^t)] 
\end{align}
where $\ell_{ce}$ is the standard cross-entropy loss.

Joint training achieves the best performance for all the tasks (in theory) as the network is trained with the data from all the tasks simultaneously, most of which is no longer accessible in lifelong learning setup.
To tackle the lack of previous data, LwF~\cite{li2017learning} suggests that we should first record the response of $C^*_t \circ C^* \circ F^*$ (i.e., classification probabilities) on $X^\tau$, and replace the $\mathcal{L}_{old}$ with:
\begin{equation}
\label{eq:dist}
\mathcal{L}_{dist} = \sum^{\tau-1}_{t=1} \mathbb{E}[\ell_{dist}(C_t \circ C \circ F(X^\tau), C^*_t \circ C^* \circ F^*(X^\tau))]
\end{equation}
where $C^*_t \circ C^* \circ F^*$ are the model parameters optimal for the old tasks and $C_t \circ C \circ F$ are those to be learned for the new task.
$\ell_{dist}$ denotes the knowledge distillation loss (KD loss) proposed in \cite{hinton2015distilling}.

However, LwF is reported to suffer an accumulating drop in performance as the task sequence grows longer~\cite{aljundi2017expert}.
EBLL~\cite{rannen2017encoder} slightly alleviates the forgetting problem with auto-encoders to preserve the necessary information of old tasks in a lower dimensional manifold.
Nevertheless, its improvement over LwF is mainly for the old tasks while the performance on the new task is even inferior~\cite{hou2018lifelong}.

\begin{figure*}
  \centering
  \includegraphics[width=0.9\linewidth]{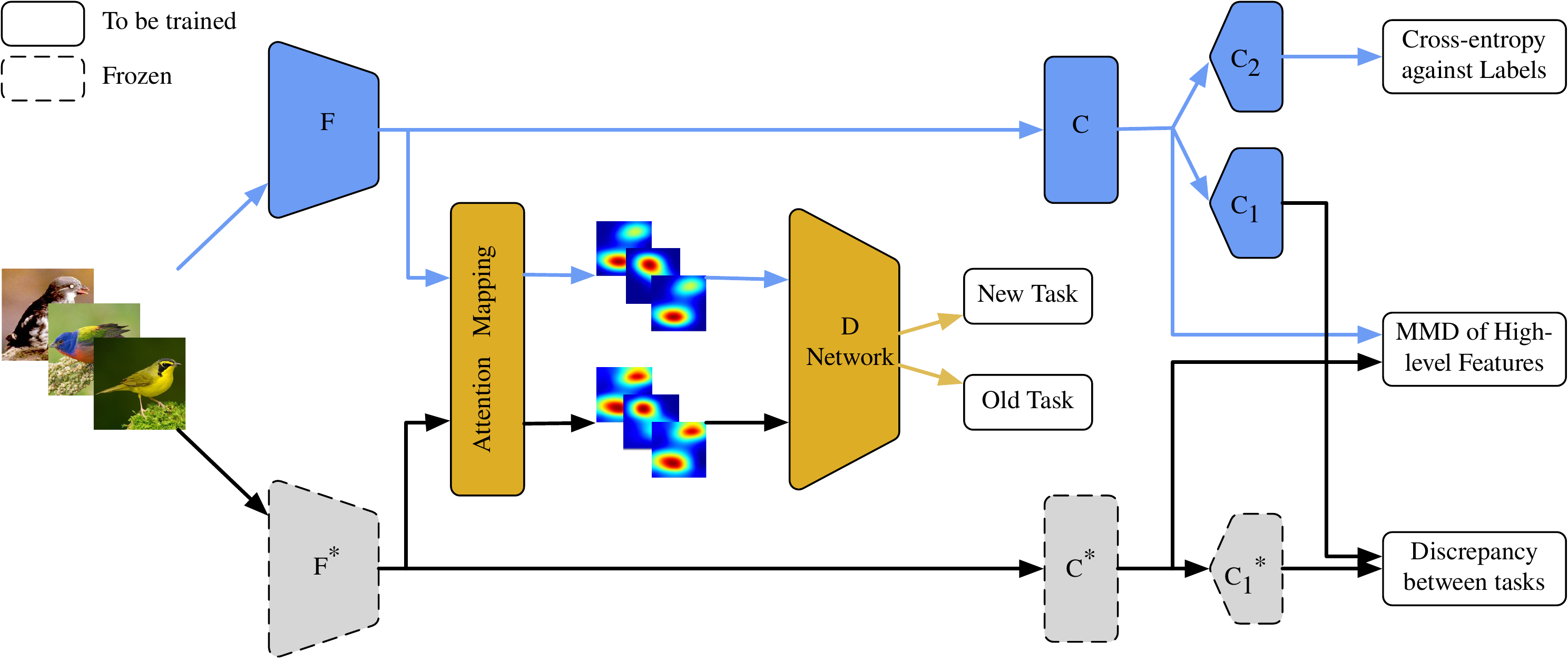}
  \caption{The architecture of the proposed method. Feature alignment penalty is introduced in addition to the cross-entropy loss against labels and the distillation loss between the label probabilities of the old and new models: the convolutional feature maps generated by different models with the same data are aligned through adversarial attention alignment (Section \ref{sec:aaa}); the high-level semantic features across networks are aligned by MMD (Section \ref{sec:mmd}). The above two constraints provide supervised information of the old task while acting as regularizers for the new tasks. (Better viewed in color)}
  \label{fig:ours}
\end{figure*}

\section{Method}
\label{sec:method}
Classification probabilities of the old model cannot provide as strong and accurate supervised information as the labels, which accounts for the performance drop in LwF and EBLL.
From another point of view, the old model $f^*$ has learned the knowledge of previous tasks.
Thus the intermediate activations (or features) of $f^*$ can be treated as soft targets and guide the training process.
Inspired these insights, we propose a novel activation regularization based lifelong learning method with adversarial feature alignment~(AFA).

Our framework is illustrated in Figure \ref{fig:ours}, with a two-stream model denoting the old and new networks respectively.
Besides the cross-entropy loss of the new task and the constraint between classification probabilities of the previous tasks, we employ the low-level visual features and high-level semantic features as soft targets, aiming to provide sufficient supervised information of the old task through multilevel feature alignment.




\subsection{Adversarial Attention Alignment}
\label{sec:aaa}

It is challenging to use convolutional visual features as soft targets, as these low-level features are in high dimensions and hard to characterize.
For example, the dimension of $conv5$ activation of AlexNet~\cite{krizhevsky2012imagenet} is 9216 ($6 \times 6 \times 256$) when flattened into a vector.
Directly putting this vector into a neural network model will bring a vast amount of parameters while using statistic moments (e.g., L2 norm) as constraints will lose the 2D structural information in the feature maps (refer to Section~\ref{sec:c_conv} for more detailed discussions).
Here, we solve the problem with the help of activation-based visual attention mechanism,
which is defined as a function of spatial maps w.r.t. the convolutional layers as in~\cite{Zagoruyko2017AT}.
The attention mechanism will put more weight on the most discriminative parts and make it easier to capture the character of visual feature maps.

Concretely, let us consider a convolutional layer and its corresponding activation tensor $A \in R^{C \times H \times W}$, which consists of $C$ channels with spatial dimensions $H \times W$.
An activation-based mapping function $F_{att}$ takes the above 3D tensor $A$ as input and outputs a spatial attention map, i.e., a flattened 2D tensor defined over the spatial dimensions:
\begin{equation}
F_{att} : R^{C \times H \times W} \to R^{H \times W}
\end{equation}

An implicit assumption to define such a spatial attention map function is that the absolute value of a hidden neuron activation can be used as an indication of the importance of that neuron w.r.t. the specific input.
Specifically, we consider the following activation-based mapping function:
\begin{equation}
\label{eq:att}
F_{att} (A) = \sum^C_{ch=1} |A_{ch}|^2,\quad A_{ch} \in R^{H \times W}
\end{equation}
where $A_{ch}$ is the $ch$-th feature map of activation tensor $A$.
The operations in Eq.(\ref{eq:att}) are element-wise.

Based on the above attention mapping, we further propose to apply the adversarial alignment penalty to the attention maps of visual features, which guides the new model to integrate the knowledge from the old model.
A discriminator (termed as $D$ network) is introduced to play the GAN-like minimax game with the feature extractors of the old and new models, i.e., $F$ and $F^*$.
Formally, the feature extractors take the training data as inputs and compute convolutional feature maps, which are further encoded by attention mapping function $F_{att}$ into latent representations $Z=\{z\}$, where:
\begin{equation}
\label{eq:z}
z = F_{att} \circ F(x),\quad x \in X^{\tau}
\end{equation}

The $D$ network tries to distinguish that the latent representations come from the old or new network.
Thus, $D$ is optimized by a standard supervised loss in GAN, defined as follow:
\begin{equation}
\label{eq:adv_d}
\mathcal{L}_{adv_D} = \max\limits_D\mathbb{E}_{z^* \sim Z^*}[\text{log}D(z^*)]+\mathbb{E}_{z \sim Z}[\text{log}(1 - D(z))]
\end{equation}
where $Z^*$ and $Z$ are latent representations from the old and new feature extractors respectively.

Then the feature extractor $F$ is updated by playing minimax game with the discriminator $D$.
Rather than directly adopting the gradient reverse layer~\cite{ganin2016domain}, which corresponds to the true minimax objective, modern GANs are usually trained with inverted label loss, defined as:
\begin{equation}
\label{eq:adv_f}
\mathcal{L}_{adv_F} = \min\limits_F-\mathbb{E}_{z \sim Z}[\text{log}D(z)]
\end{equation}
This objective shares the same convergence with the minimax loss but provides stronger gradients and thus eases the training process.

\textbf{Discussion:}
In our method, discriminator $D$ plays the minimax game with feature extractor $F$, instead of a generator in standard GANs.
Similar modules are commonly adopted in adversarial domain adaptation studies~\cite{ganin2016domain,tzeng2017adversarial, kang2018deep}.
The significant difference between them and our method lies in the dataflow.
We use one single dataflow, freeze the old model $f^*$ and align features generated by different models yet with the same data to distill the knowledge of the old network to the new one, instead of pairing features generated from different data distributions in domain adaptation works.

\subsection{High-level Feature Alignment with MMD}
\label{sec:mmd}

In deep CNNs, the features generated by fully connected layers are high-level semantic representations, which contain massive information about tasks and labels.
Aligning these high-level task-specific features will force the new model to integrate the knowledge about the old tasks learned by the old network.
However, employing a discriminator here and playing minimax game with $C \circ F$ does not work, because these modules have already been customized for specific tasks and thus cannot adapt immediately to confuse the $D$ network 
(refer to Section~\ref{sec:c_fc} for more detailed discussions).

Previous studies on measuring the discrepancy between high-level features usually take advantage of MMD~\cite{gretton2012optimal}.
Concretely, given two data distributions $P$ and $Q$, MMD is expressed as the distance between their means after mapped to a reproducing kernel Hilbert space (RKHS):
\begin{equation}
\label{eq:mmd1}
\mathit{MMD}^2(P, Q) = \| \mathbb{E}_{p \sim P}[\phi(p)] - \mathbb{E}_{q \sim Q}[\phi(q)] \|^2
\end{equation}
where $\phi(\cdot)$ denotes the mapping to RKHS.

In practice, this mapping is unknown.
Expanding Eq. (\ref{eq:mmd1}) and using the kernel trick to replace the inner product, we have an unbiased estimator of MMD:
\begin{equation}
\label{eq:mmd2}
\mathcal{L}_{mmd}(P, Q) = \mathbb{E}_{p,q\sim P, Q}[k(p, p) + k(q, q) - 2k(p, q)]
\end{equation}
where $k(p, q) = \langle\phi(p), \phi(q)\rangle$ is the desired kernel function.
In this work, we use a standard radial basis function (RBF) kernel with multiple widths~\cite{gretton2012optimal}.

Formally, let $H = \{h\}$ be the fully connected features generated by the feature extractor $F$ and the shared classifier $C$:
\begin{equation}
\label{eq:h}
h = C \circ F(x),\quad x\in X^\tau
\end{equation}

We align the high-level semantic features from the old and new models (termed as $H^*$ and $H$ respectively) by minimizing the following loss function:
\begin{equation}
\label{eq:fc}
\mathcal{L}_{fc} = \mathcal{L}_{mmd}(H, H^*)
\end{equation}

\textbf{Discussion:}
In previous studies, MMD is usually used for facilitating the network to generate domain-invariant features for data from the same class but different domains.
In our method, the fully connected features are generated by different networks but with the same data.
They contain rich task information and can be treated as soft targets for the new model.
By aligning these high-level semantic features, the knowledge of previous tasks is transferred across networks.

\subsection{Overall Algorithm}


The backbone network is trained by minimizing the following loss function that consists of four parts of constraints:
\begin{align}
\label{eq:ours}
\mathcal{L} &= \mathcal{L}_{cls} + \lambda_1\mathcal{L}_{dist} + \lambda_2\mathcal{L}_{adv_F} + \lambda_3\mathcal{L}_{fc}
\end{align}
where $\mathcal{L}_{cls}$, $\mathcal{L}_{dist}$, $\mathcal{L}_{adv_F}$ and $\mathcal{L}_{fc}$ are defined in Eq.(\ref{eq:cls}), (\ref{eq:dist}), (\ref{eq:adv_f}) and (\ref{eq:fc}).
$\lambda_1$, $\lambda_2$ and $\lambda_3$ are hyperparameters, of which some discussions are available in Section \ref{sec:hyper_param}.

The key idea of our method lies in employing the intermediate activations of the old model, which contains rich knowledge of previous tasks, as soft targets to guide the training process in multiple stages when fitting the new data.
Aligning the multilevel features provides sufficient supervised information of the old tasks and helps to alleviate the forgetting problem.
Further experiments demonstrate that the feature alignment strategy enables the model to gain even better performance than finetuning on the new tasks due to the knowledge distillation and regularization phenomenons.

\section{Experiments}

Our method is compared with the state-of-the-art regularization-based lifelong learning methods and several baselines in the incremental task scenarios.
We consider the situations containing two (starting from large and small datasets separately) and five tasks.

\subsection{Architecture}

The network architecture is base on the AlexNet~\cite{krizhevsky2012imagenet}, which is a representative of CNNs and widely used in transfer learning research~\cite{yosinski2014transferable,long2015learning,li2017learning,rannen2017encoder,aljundi2018memory}.
Transferability studied on it can be extended to other network architectures easily.
Concretely, the shared feature extractor $F$ corresponds to the convolutional layers, i.e., $\mathit{conv1\sim conv5}$.
The shared classifier $C$ denotes all the fully connected layers except the last one, i.e., $\mathit{fc6}$ and $\mathit{fc7}$, while the task-specific classifier is the last fully connected layer $\mathit{fc8}$.
The network architecture of our method is illustrated in Figure \ref{fig:ours}.

\subsection{Datasets}

We use the popular datasets in the incremental task scenario~\cite{li2017learning,aljundi2017expert,rannen2017encoder,aljundi2018memory}, including:
\begin{itemize}
\item MIT Scenes~\cite{quattoni2009recognizing}: images for indoor scenes classification, including 5360 training samples and 1340 test samples.
\item Caltech-UCSD Birds~\cite{WelinderEtal2010}: images for fine-grained bird classification, including 5994 training samples and 5794 test samples.
\item Oxford Flowers~\cite{Nilsback08}: images for fine-grained flower classification, including 2040 training samples and 6149 test samples.
\item Stanford Cars~\cite{KrauseStarkDengFei-Fei_3DRR2013}: images for car classification, including 8,144 training samples and 8,041 testing samples.
\item FGVC-Aircraft~\cite{maji13fine-grained}: images for aircraft manufacturer classification, including 6667 training samples and 3333 testing samples.
\item ImageNet (ILSVRC 2012 subset)~\cite{deng2009imagenet}: the validation set of ILSVRC 2012 is used for testing the effect of performance preservation on old tasks.
\end{itemize}

The results reported in this paper are obtained on the test sets of Scenes, Birds, Flowers, Cars and Aircraft, and the validation set of ImageNet.
We \emph{fix the rand seed} and make the results reproducible.

\subsection{Compared Methods}

The proposed \textit{AFA} is compared with the following methods:
\begin{itemize}
\item \textit{Joint Training}: The data of all the tasks are used during training, which is considered as an upper bound of performance preservation on old tasks.
\item \textit{Finetuning}: Copy the feature extractor ($F$) and shared classifier ($C$) of the old model, randomly initialize the task-specific classifier ($C_\tau$) and then train the network on the new task $\tau$.
\item \textit{LwF}~\cite{li2017learning}: They introduce a KD loss term between the label probabilities of the old and new models computed on the new data to preserve the knowledge of previous tasks.
\item \textit{EBLL}~\cite{rannen2017encoder}: This work builds on LwF and additionally prevents the reconstructions of convolutional features from changing with auto-encoders to reduce forgetting.
\item \textit{EWC}~\cite{kirkpatrick2017overcoming}: It estimates the importance of parameters by the diagonal of the Fisher information matrix and applies a penalty to the change of important parameters during training on the new task. It uses individual penalty for each previous task.
\item \textit{SI}~\cite{zenke2017continual}: It is similar to EWC but estimates the importance weights in an online manner by the parameter specific contribution to the changes in the total loss.
\item \textit{MAS}~\cite{aljundi2018memory}: It is similar to SI but estimates the importance weights by the changes of the model outputs.
\end{itemize}

As stated before, we do not consider lifelong learning methods that require storing model parameters (i.e., architectural strategies) or training samples (i.e., rehearsal strategies).

\textbf{Remark:}
The original versions of parameter regularization strategies (e.g., \textit{EWC} and \textit{SI}) require the network to maintain the same structure throughout all tasks as the constraint is explicitly given to each parameter.
To make these methods compatible with incremental task scenario where tasks have different output domains, we apply the parameter constraints to feature extractor ($F$) and shared classifier ($C$), excluding task-specific classifiers ($C_{1,...,\tau}$).
The results are reported after convergence on new tasks.

\begin{table*}
\caption{Classification accuracy in the two-task scenario starting from \emph{ImageNet}}
\label{table:single}
\centering
\begin{threeparttable}
\begin{adjustbox}{width=\textwidth}
\begin{tabular}{lllllllll}
\toprule
\multirow{2}{*}{}
&\multicolumn{2}{c}{ImageNet$\to$Scenes}
&\multicolumn{2}{c}{ImageNet$\to$Birds}
&\multicolumn{2}{c}{ImageNet$\to$Flowers}
&\multicolumn{2}{c}{Average} \\
\cmidrule(l){2-3}
\cmidrule(l){4-5}
\cmidrule(l){6-7}
\cmidrule(l){8-9}
&\multicolumn{1}{c}{Old Task}&\multicolumn{1}{c}{New Task}
&\multicolumn{1}{c}{Old Task}&\multicolumn{1}{c}{New Task}
&\multicolumn{1}{c}{Old Task}&\multicolumn{1}{c}{New Task}
&\multicolumn{1}{c}{Old Task}&\multicolumn{1}{c}{New Task}\\
\midrule
Joint Training~\tnote{1}
& 55.11\ (ref) & 62.93\ (+0.14)
& 54.93\ (ref) & 56.88\ (-0.29)
& 56.26\ (ref) & 85.09\ (-0.21)
& (ref) & (-0.12)\\

Finetuning~\tnote{2}
& 51.28\ (-3.83)  & 62.79\ (ref)
& 42.94\ (-11.99) & 57.17\ (ref)
& 44.46\ (-11.80) & 85.30\ (ref)
& (-9.21) & (ref)\\
\midrule

LwF
& 53.62\ (-1.49) & 63.51\ (+0.72)
& 53.41\ (-1.52) & 57.42\ (+0.25)
& 54.64\ (-1.62) & 85.15\ (-0.15)
& (-1.54) & (+0.27)\\

EBLL
& 54.33\ (-0.78) & 63.29\ (+0.50)
& 54.17\ (-0.76) & 56.78\ (-0.39)
& \textbf{55.37\ (-0.89)} & 84.36\ (-0.94)
& (-0.81) & (-0.28)\\
\midrule

EWC
& 54.25\ (-0.86) & 61.34\ (-1.45)
& 52.16\ (-2.77) & 54.57\ (-2.60)
& 53.81\ (-2.45) & 84.58\ (-0.72)
& (-2.03) & (-1.59)\\

MAS
& 53.98\ (-1.13) & 61.87\ (-0.92)
& 53.06\ (-1.87) & 54.26\ (-2.91)
& 54.90\ (-1.36) & 83.66\ (-1.64)
& (-1.46) & (-1.83)\\
\midrule

\textbf{AFA}
& \textbf{54.71\ (-0.40)} & \textbf{63.88\ (+1.09)}
& \textbf{54.43\ (-0.50)} & \textbf{57.84\ (+0.66)}
& 55.21\ (-1.05) & \textbf{86.03\ (+0.73)}
& \textbf{(-0.65)} & \textbf{(+0.83)}\\
\bottomrule
\end{tabular}
\end{adjustbox}
\begin{tablenotes}
    \item[1] For the old task, the reference performance is given by \emph{Joint Training}.
    \item[2] For the new task, \emph{Finetuning} is considered as the reference.
\end{tablenotes}
\end{threeparttable}
\end{table*}

\subsection{Implementation Details}
\label{sec:imple}

During training, we use the stochastic gradient descent (SGD) optimizer with a momentum~\cite{sutskever2013importance} of 0.9 and dropout enabled in the fully connected layers.
The data normalization is applied within each task.
We augment the training data with random resized cropping and random horizontal flipping but without color jittering.

We adopt a three-layer perceptron as the discriminator, with the hidden layer consisting of 500 units.
Additionally, we find that the number of neurons in the hidden layer is not a sensitive hyperparameter.
Choices in the range from 100 to 900 will produce very close results.

Before training, we randomly initialize $C_\tau$ using the Xavier~\cite{glorot2010understanding} initialization with a scaling factor of 0.25.
Then we freeze $F$ and $C$, and train $C_\tau$ for some epochs, which is termed as the \emph{warm-up} step in LwF.
In our experiments, the warm-up step lasts for 70 epochs, and then the whole network is trained until convergence.
For a fair comparison, the warm-up model is used as the start point for all the compared methods.
When convergence on the validation set is observed, we reduce the learning rate by $0.1\times$ and keep training for extra 20 epochs.

\subsection{Two-Task Scenario Starting from \emph{ImageNet}}

In the two-task scenario, we are given a network trained on a previous task, and then one single new task arrives to be learned.
Firstly, we follow the experimental setup in \textit{LwF} and \textit{EBLL}, in which all the experiments start from an AlexNet model pre-trained on ImageNet.
This scenario contains three individual experiments, i.e., ImageNet $\to$ Scenes, ImageNet$\to$Birds, and ImageNet$\to$Flowers.
The performances of our method and compared ones in the two-task scenario are illustrated in Table \ref{table:single}.
SI~\cite{zenke2017continual} is not included in this scenario as it requires training from scratch on ImageNet to get the importance weights.

As Table \ref{table:single} shows, the reference performance on the old task is given by \textit{Joint Training}, which assumes that the data of previous tasks are available.
Without any constraints, \textit{Finetuning} favors the new task but does not care about the performance on the old tasks as the training process goes forward.
The performances on ImageNet drop dramatically, especially in ImageNet$\to$Birds and ImageNet$\to$Flowers.
That is exactly the \emph{Catastrophic Forgetting} problem that we are trying to overcome.
Through activation or parameter regularization, \textit{LwF}, \textit{EBLL}, \textit{EWC} and \textit{MAS} alleviate the forgetting problem to some extent.
Our method \textit{AFA} suffers the least drop in performance on the old task in most of the three experiments.
The visual and semantic features generated by the old model contain rich knowledge of the old task, which is integrated into the new model through the proposed feature alignment strategy.

Considering the performance on the new task, \textit{AFA} reaches the best performance among the compared methods, even better than \textit{Joint Training}, which has access to the max amount of data, and \textit{Finetuning}, a commonly used transfer learning routine.
Since the new tasks in our experiments are small datasets compared to ImageNet, \textit{Finetuning} with such data usually overfits the training sets and cannot reach high accuracy on the test sets.
\textit{AFA} and \textit{LwF} outperform \textit{Joint Training}, which is unexpected but does make sense.
The critical point is that the quantities of data from the old and new tasks are incredibly unbalanced.
Hence the network favors the data from ImageNet but ignores those of Scenes/Birds/Flowers during the training process of \textit{Joint Training}.
On the contrary, \textit{AFA} and \textit{LwF} use knowledge from both the old and new tasks and avoid the unbalanced data distribution problem, which accounts for the improvement of accuracy on the new task.
\textit{EBLL}, \textit{EWC} and \textit{MAS} focus more on preserving the performance on the old task while the accuracy on the new task is inferior.
According to \cite{kirkpatrick2017overcoming}, the layers closer to the output are indeed being reused in \textit{EWC}.
However, when the tasks have different output domains, the constraints of parameters near the output layer will prevent the model from achieving competitive performance on the new task.

\begin{table*}
\caption{Classification accuracy in the two-task scenario starting from \emph{Flowers}}
\label{table:flowers}
\centering
\begin{threeparttable}
\begin{adjustbox}{width=0.87\textwidth}
\begin{tabular}{lllllll}
\toprule
& \multicolumn{2}{c}{Flowers to Scenes}
& \multicolumn{2}{c}{Flowers to Birds}
& \multicolumn{2}{c}{Average}\\
\cmidrule(l){2-3}
\cmidrule(l){4-5}
\cmidrule(l){6-7}
& \multicolumn{1}{c}{Old Task} & \multicolumn{1}{c}{New Task}
& \multicolumn{1}{c}{Old Task} & \multicolumn{1}{c}{New Task}
& \multicolumn{1}{c}{Old Task} & \multicolumn{1}{c}{New Task} \\
\midrule
Joint Training~\tnote{1}
& 84.75\ (ref) & 61.05\ (0.67)
& 83.04\ (ref) & 56.42\ (0.22)
& 83.89\ (ref) & 58.73\ (0.45)\\

Finetuning~\tnote{2}
& 71.67\ (-13.08) & 60.37\ (ref)
& 65.41\ (-17.63) & 56.20\ (ref)
& 68.54\ (-15.35) & 58.28\ (ref)\\
\midrule

LwF
& 79.58\ (-5.16) & 62.39\ (2.02)
& 79.35\ (-3.69) & 55.89\ (-0.31)
& 79.46\ (-4.43) & 59.14\ (0.85)\\

EBLL
& 80.19\ (-4.56) & 61.57\ (1.20 )
& 80.09\ (-2.95) & 55.26\ (-0.94)
& 80.14\ (-3.75) & 58.42\ (0.13)\\
\midrule

EWC
& 78.89\ (-5.85) & 58.96\ (-1.42)
& 76.57\ (-6.47) & 55.07\ (-1.12)
& 77.73\ (-6.16) & 57.01\ (-1.27)\\

SI
& 78.78\ (-5.97) & 58.88\ (-1.49)
& 76.50\ (-6.54) & 55.13\ (-1.07)
& 77.64\ (-6.25) & 57.00\ (-1.28)\\

MAS
& 78.89\ (-5.85) & 58.51\ (-1.87)
& 76.63\ (-6.41) & 55.06\ (-1.14)
& 77.76\ (-6.13) & 56.78\ (-1.50)\\
\midrule

\textbf{AFA}
& \textbf{80.96\ (-3.78)} & \textbf{63.73\ (3.36)}
& \textbf{81.15\ (-1.88)} & \textbf{57.60\ (1.40)}
& \textbf{81.06\ (-2.83)} & \textbf{60.66\ (2.38)} \\
\bottomrule
\end{tabular}
\end{adjustbox}
\begin{tablenotes}
    \item[1] For the old task, the reference performance is given by \emph{Joint Training}.
    \item[2] For the new task, \emph{Finetuning} is considered as the reference.
\end{tablenotes}
\end{threeparttable}
\end{table*}

\subsection{Two-Task Scenario Starting from \emph{Flowers}}

When we start from a smaller dataset, \textit{Flowers} here, different trends can be observed.
The experimental results under Flowers $\to$ Scenes and Flowers$\to$Birds are illustrated in Table \ref{table:flowers}.
Similar to that starting from ImageNet, the reference performance of the old task is given by \textit{Joint Training} while \textit{Finetuning} is considered as the reference for the new task.

Activation regularization strategies (\textit{LwF}, \textit{EBLL} and \textit{AFA}) outperform parameter regularization strategies (\textit{EWC}, \textit{SI} and \textit{MAS}) in both alleviating forgetting on the old task and performance on the new task.
One potential reason is due to the biased estimation of importance weights.
The data samples are too few to compute the importance weights w.r.t. the old tasks when starting from small datasets.
Another side effect of the biased importance weights is the poor performances on the new tasks.

Among the activation regularization strategies, the improvement brought by \textit{EBLL} compared to \textit{LwF} is mainly for the performance preservation on old tasks while the accuracy on the new task is inferior.
The conclusion is consistent with that in PDG~\cite{hou2018lifelong}.
Under the guidance of multilevel soft targets, the proposed method \textit{AFA} further alleviates the forgetting problem compared to \textit{LwF} and \textit{EBLL}.
Meanwhile, \textit{AFA} achieve a better performance than \textit{Finetuning} and \textit{Joint Training} on the new task due to the regularization and knowledge distillation phenomenons.

\textbf{In conclusion}, the proposed \textit{AFA} suffers the \emph{least performance drop on the old task} while achieving the \emph{best accuracy on the new task} among all the compared methods in scenarios starting from both large (e.g., ImageNet) and small (e.g., Flowers) datasets.

\subsection{Five-Task Scenario}

\begin{figure}[t]
    \centering
    \includegraphics[width=\linewidth]{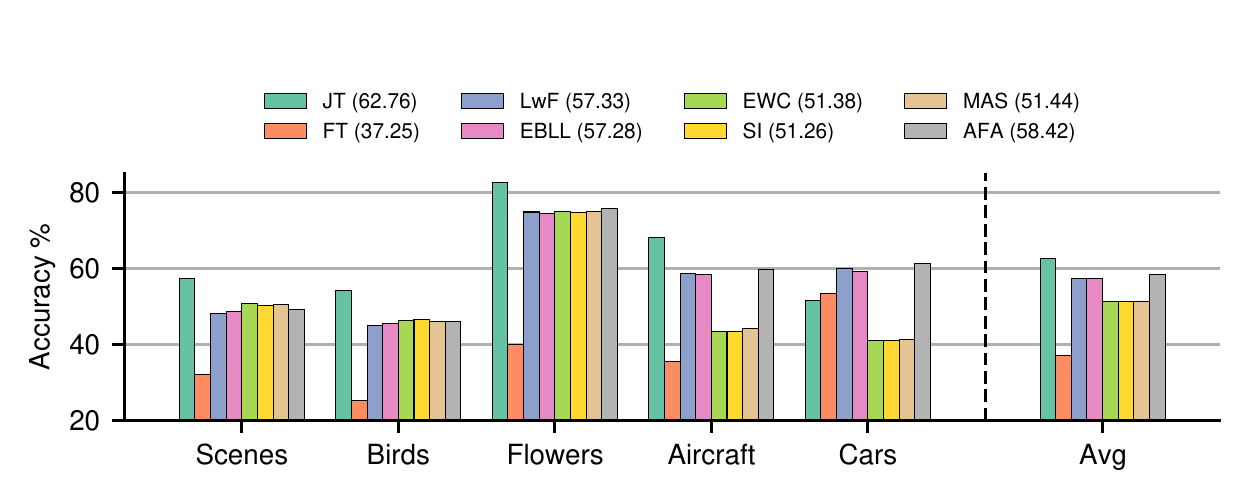}
    \caption{The performance on each task at the end of the five-task sequence. The average accuracy over five tasks is marked next to the legend. (Better viewed in color)}
    \label{fig:5tasks}
\end{figure}

We further study a more challenging scenario containing a sequence of five tasks: Scenes$\to$Birds$\to$Flowers$\to$Aircraft$\to$Cars.
The performances of all compared methods on each task at the end of the five-task sequence are illustrated in Figure \ref{fig:5tasks}.
For a more efficient training process, we start from the AlexNet model pre-trained on ImageNet and then finetuned on Scenes.
We do not include the performance on ImageNet in the results as \textit{SI}~\cite{zenke2017continual} requires training from scratch on ImageNet to get the importance weights.

As expected, \textit{Finetuning} suffers severe forgetting on previous tasks and favors the most recently learned task.
With access to all the training data, \textit{Joint Training} reaches the highest accuracy on the first four tasks and the best performance on average.

Parameter regularization strategies (\textit{EWC}, \textit{SI}, and \textit{MAS}) better preserve the performance on older tasks and outperform activation regularization strategies in the first two tasks Scenes and Birds.
Nevertheless, as has been stated many times, parameter regularization strategies usually prevent the model from achieving competitive performance on the new tasks, which accounts for the reduced accuracy in the last two tasks Aircraft and Cars.

Recall that in the two-task scenario, both reported in our experiments and original paper~\cite{li2017learning}, \textit{LwF} reaches a higher accuracy than \textit{Joint Training} on the new task.
However, in the five-task scenario, the result is reversed, which is consistent with that discussed before: \textit{LwF} suffers an accumulating drop in performance as the sequence grows longer.
\textit{EBLL} brings some improvements in alleviating the forgetting problem compared to \textit{LwF}, but the performances on newer tasks are inferior.
\textit{EBLL} operates on the high dimensional features generated by the convolutional layers directly and loses the structural information of feature maps.
On the contrary, the proposed method \textit{AFA} is more effective, which distills the knowledge of previous tasks to the new model by aligning multilevel soft targets and acts as a regularizer for the new task to improve the training effect.
It reaches the best average performance with a balance between performance preservation on old tasks and accuracy on new tasks.

\begin{figure}[t]
    \centering
    \begin{subfigure}{0.495\linewidth}
        \centering
        \includegraphics[width=\linewidth]{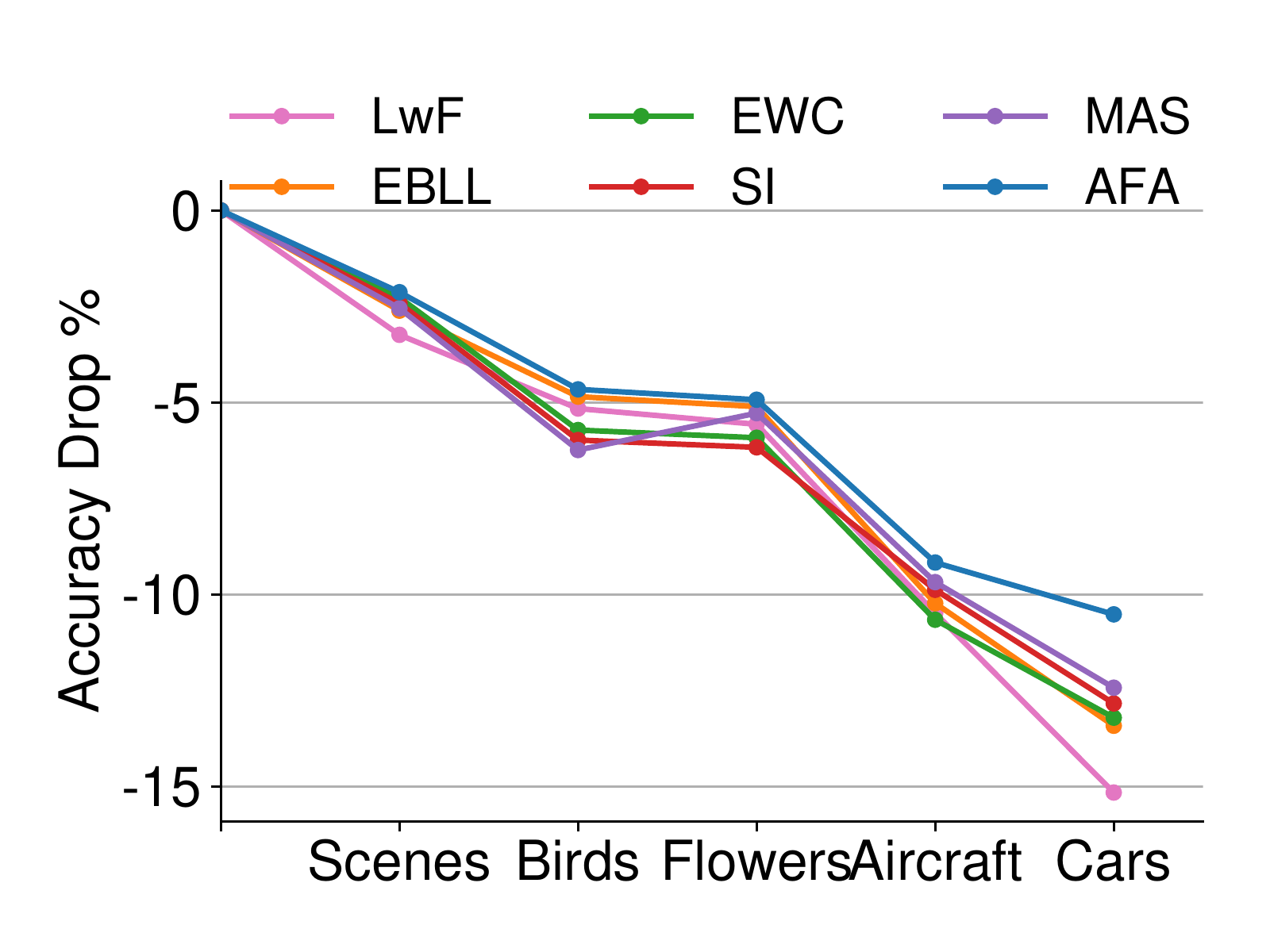}
        \caption{}
        \label{fig:gain_drop_a}
    \end{subfigure}
    \begin{subfigure}{0.495\linewidth}
        \centering
        \includegraphics[width=\linewidth]{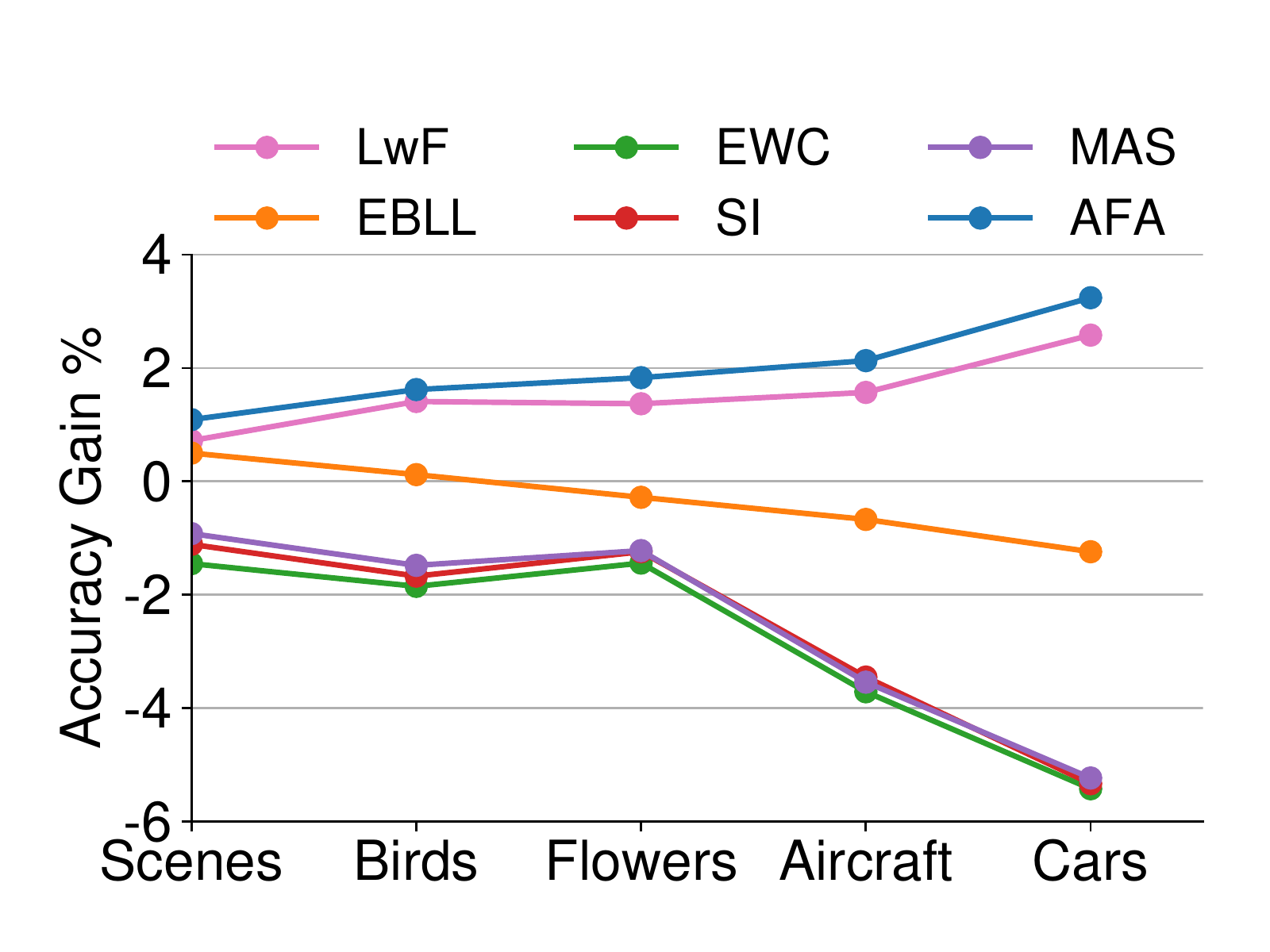}
        \caption{}
        \label{fig:gain_drop_b}
    \end{subfigure}
    \caption{(a): The average performance drop on previous tasks (relative to the performance right after training on that task) after training on a new task in five-task scenario. (b): The average performance gain on new tasks compared to \textit{Finetuning} in five-task scenario. (Better viewed in color)}
    \label{fig:gain_drop}
\end{figure}

For more detailed comparisons, we further illustrate additional information in Figure~\ref{fig:gain_drop}.
Concretely, Figure \ref{fig:gain_drop_a} shows the average performance drop over previous tasks (relative to the performance right after training on that task) after training on a new task.
Figure \ref{fig:gain_drop_b} demonstrates the average performance gain on new tasks compared to \textit{Finetuning}.

As shown in Figure \ref{fig:gain_drop_a}, the performance drop of different methods share the same trends.
The proposed \textit{AFA} suffers the least degradation in average among the compared methods.
It is worth noting that although the relative performance drop of parameter regularization strategies is comparable to \textit{AFA}, the absolute accuracy is much lower due to their poor performance on the newer tasks.
In other words, they have a lower baseline.
When it comes to new tasks, different trends can be observed.
\textit{AFA} and \textit{LwF} gain a higher accuracy than \textit{Finetuning} while \textit{EBLL} and parameter regularization strategies (\textit{EWC}, \textit{SI}, and \textit{MAS}) perform worse and worse as the training process goes forward.

\textbf{The conclusion} is identical to that in the two-task scenario.
Under the guidance of the sufficient supervised information provided by multilevel soft targets, our method \textit{AFA} alleviates the forgetting problem on old tasks while achieving even better performance than \textit{Finetuning} on new tasks.

\subsection{Ablation Study}
\label{sec:ablation}

\begin{figure}
    \centering
    \includegraphics[width=0.9\linewidth]{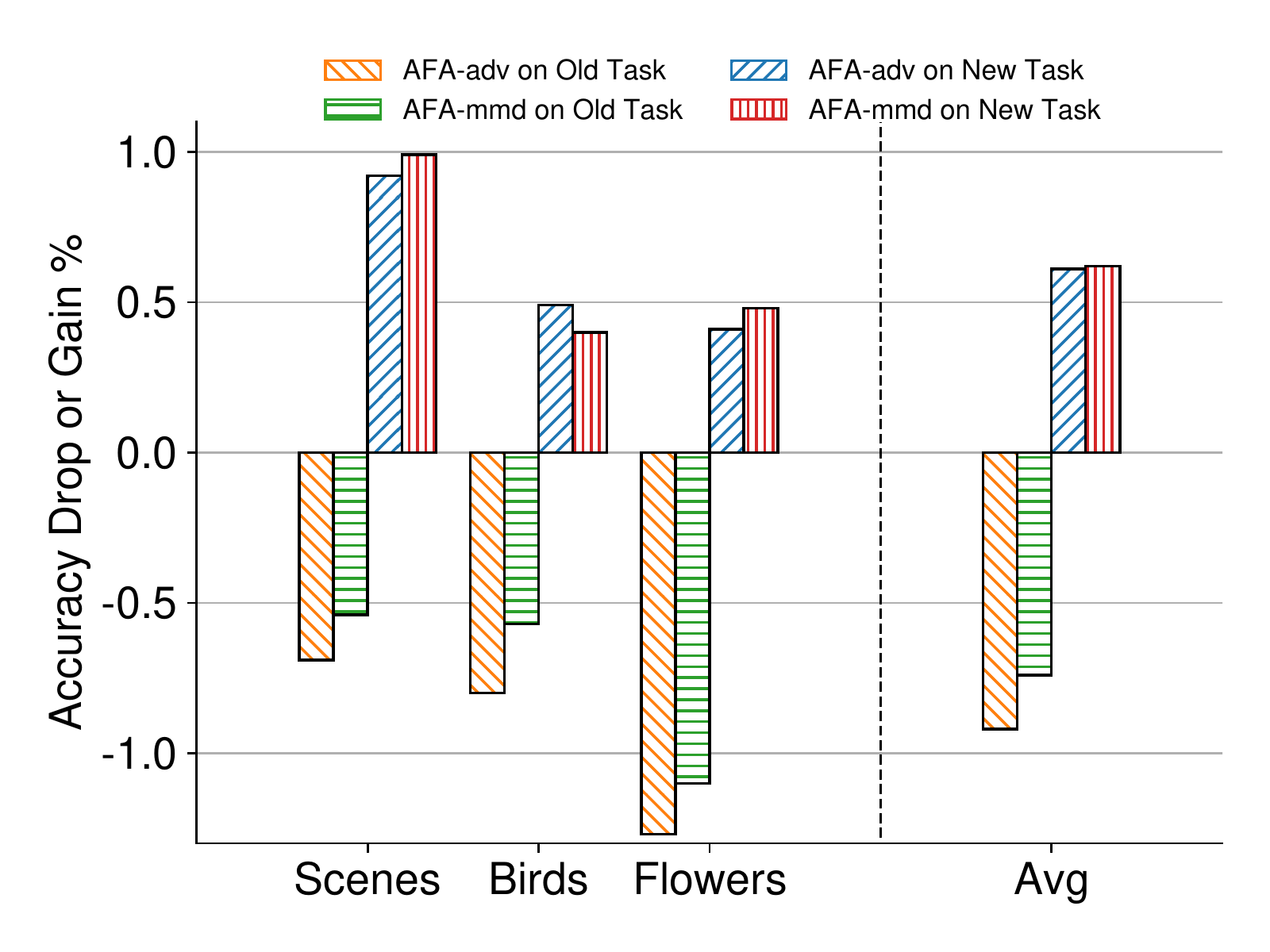}
    \caption{AFA-adv: method with merely adversarial attention alignment of visual features; AFA-mmd: method employing MMD constraints to high-level semantic features.The accuracy drop on ImageNet or gain on new tasks are illustrated. (Better viewed in color)}
    \label{fig:advmmd}
\end{figure}

\subsubsection{With single constraint}

Since two additional constraints are introduced in this work, we will analyze them individually here.
The method with only adversarial attention alignment of visual features is termed as \textit{AFA-adv} while that employing high-level feature alignment with MMD is termed as \textit{AFA-mmd}.
Their experimental results in the two-task scenario starting from ImageNet are illustrated in Figure \ref{fig:advmmd}.
As we can see, both constraints help promote the accuracy on the new task and preserve performance on the old task.

Concretely, \textit{AFA-adv} and \textit{AFA-mmd} have similar effects on improving the accuracy on the new task.
\textit{AFA-adv} performs better in ImageNet$\to$Birds while \textit{AFA-mmd} reaches higher accuracy in ImageNet $\to$ Scenes and ImageNet$\to$Flowers.
Feature alignments act as regularizers during the training process, reduce overfitting on the new data and thus improve the test accuracy.

\textit{AFA-mmd} outperforms \textit{AFA-adv} in preserving old task's performance.
The fully connected features contain rich task-specific knowledge.
Aligning these high-level semantic features with MMD will provide strong supervised information and force the network to integrate the knowledge from the old model, and thus help to alleviate the forgetting problem.


\subsubsection{Loss function between label probabilities}
\label{sec:dist}

A popular choice for measuring the discrepancy between the outputs of the old and new models is the KD loss.
As stated in~\cite{hinton2015distilling, li2017learning}, other types of constraints also work.
Through experiments, we find that L2 norm achieves better performance preservation on the old tasks while KD loss favors the new tasks.
Since we set the hyperparameter $\lambda_1$ in Eq.(\ref{eq:ours}) to 1.0 for KD loss, a smaller one should be chosen if the L2 norm is adopted as the loss function between label probabilities.

\begin{figure}[t]
    \centering
    \begin{subfigure}{0.495\linewidth}
        \centering
        \includegraphics[width=\linewidth]{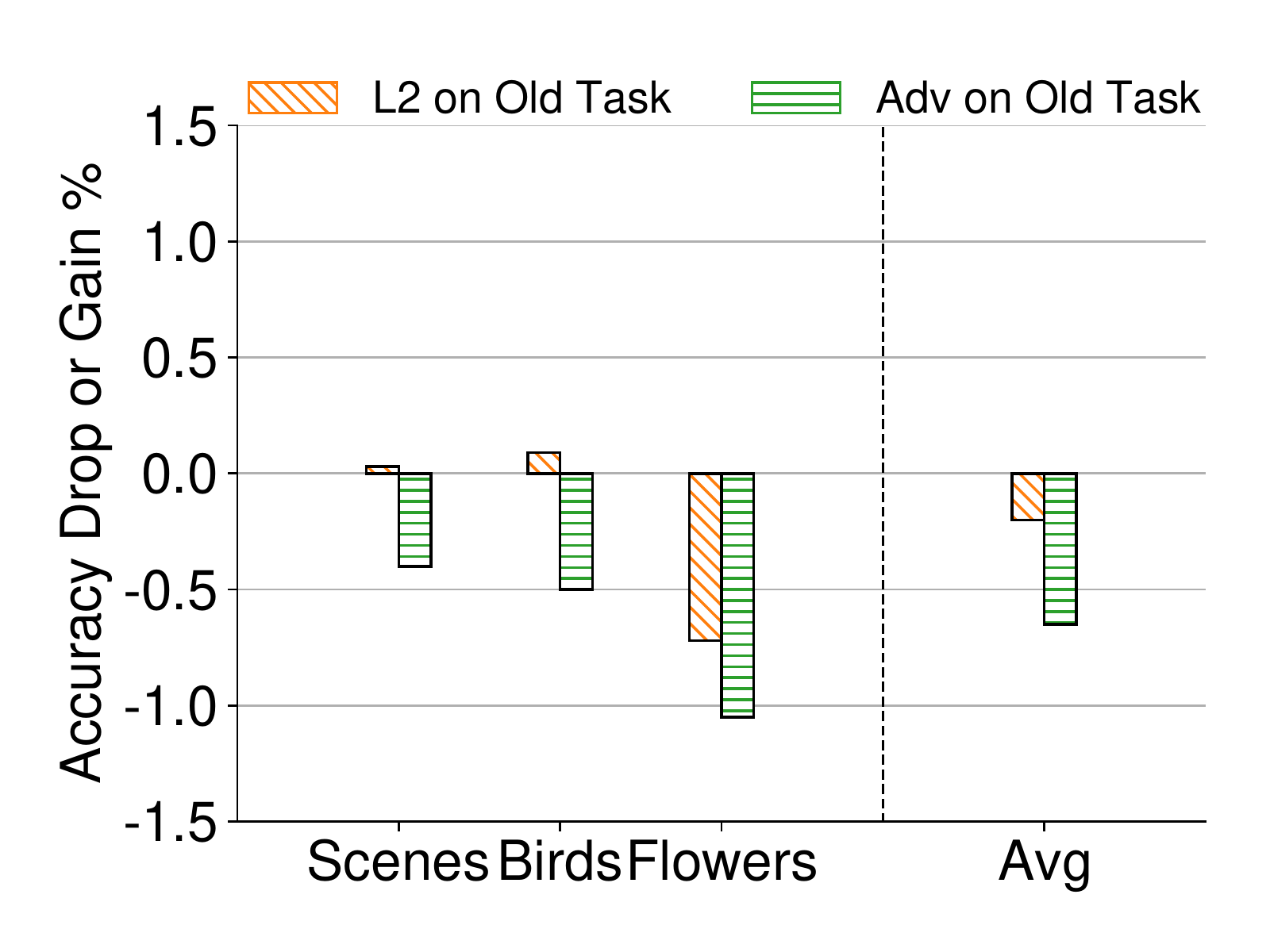}
        \caption{}
        \label{fig:conv_advl2_a}
    \end{subfigure}
    \begin{subfigure}{0.495\linewidth}
        \centering
        \includegraphics[width=\linewidth]{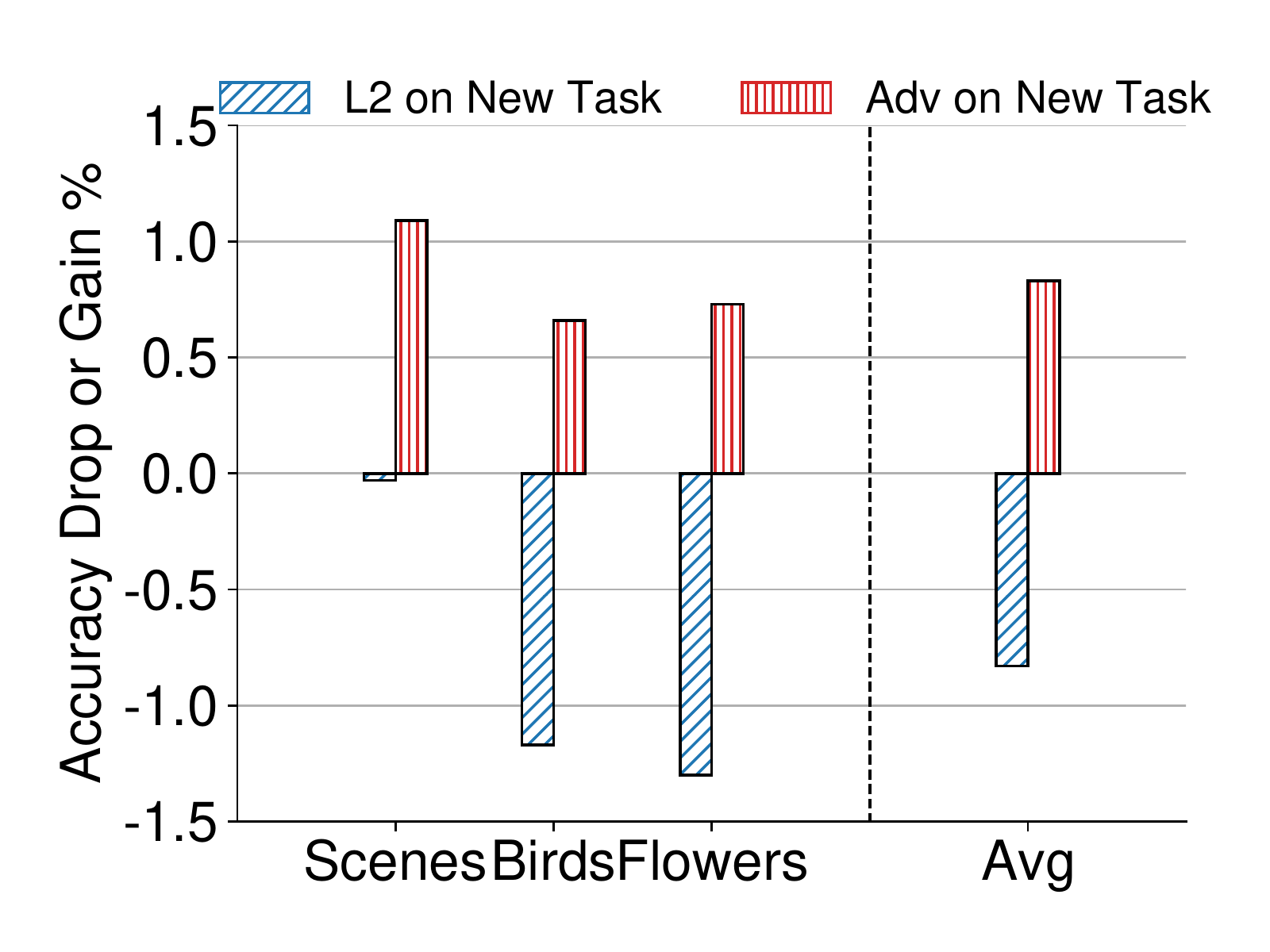}
        \caption{}
        \label{fig:conv_advl2_b}
    \end{subfigure}
    \caption{Employing \emph{Adv} or \emph{L2} as the constraints of visual features in the two-task scenario starting from ImageNet. (a): The performance drop or gain compared to \textit{Joint Training} on the old tasks; (b): The performance drop or gain compared to \textit{Finetuning} on the new tasks.}
    \label{fig:conv_advl2}
\end{figure}

\subsubsection{Constraints of visual features}
\label{sec:c_conv}

We have tried applying statistic moments (e.g., L2 norm) to the visual features directly, and the results are illustrated in Figure \ref{fig:conv_advl2}.
The method employing L2 norm achieves slightly better performance preservation on the old tasks but gets much worse accuracy on the new tasks.
Since L2 norm is a much more strict and inflexible constraint, using it as the constraint for visual features will lose the rich 2D structural information in the convolutional feature maps and prevent the model from achieving competitive performance on the new tasks.
On the contrary, the proposed adversarial attention alignment strategy introduces a trainable discriminator to measure the discrepancy between visual features dynamically and make the training process smoother.

\begin{figure}[t]
    \centering
    \begin{subfigure}{0.495\linewidth}
        \centering
        \includegraphics[width=\linewidth]{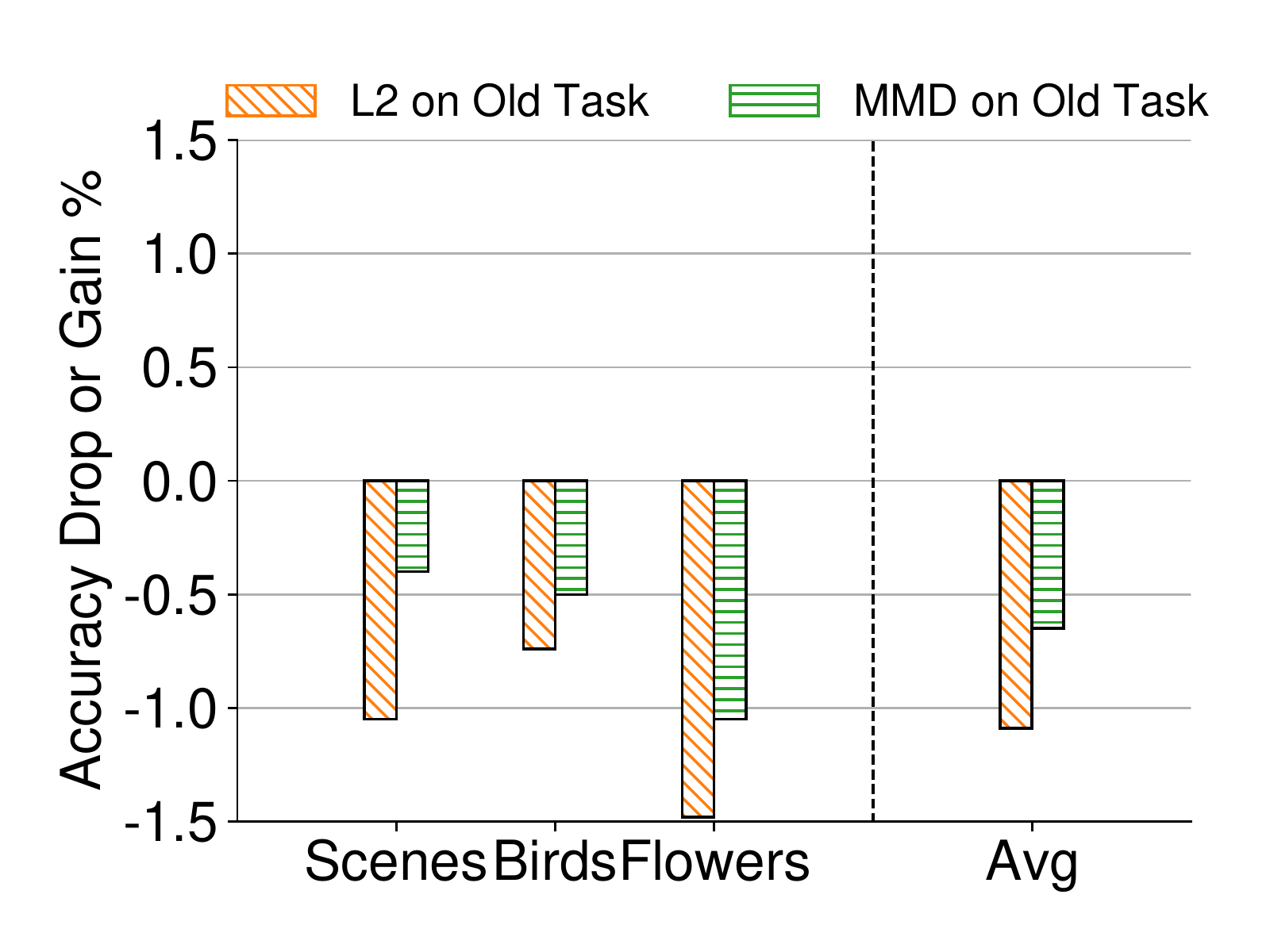}
        \caption{}
        \label{fig:fc_l2mmd_a}
    \end{subfigure}
    \begin{subfigure}{0.495\linewidth}
        \centering
        \includegraphics[width=\linewidth]{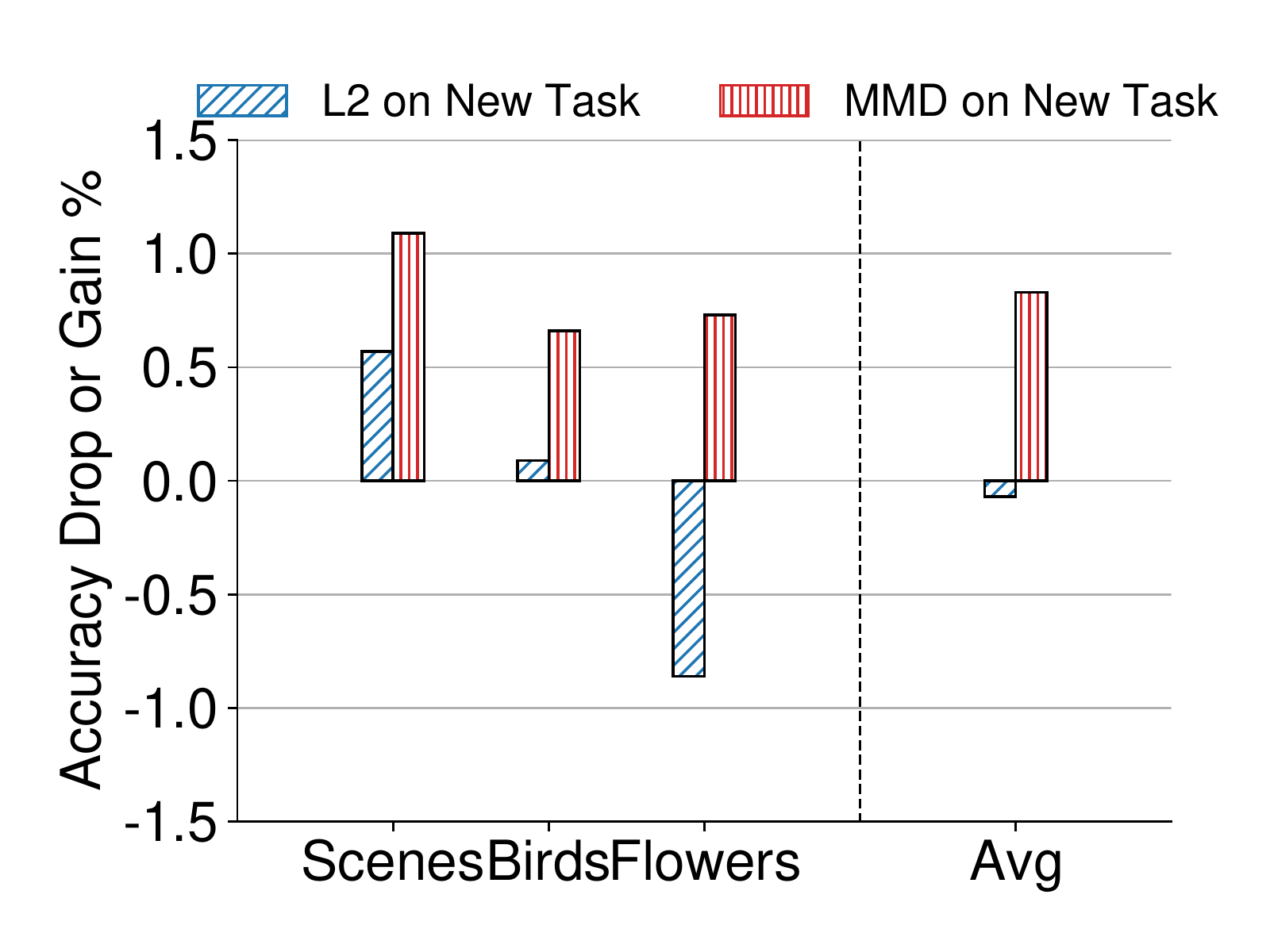}
        \caption{}
        \label{fig:fc_l2mmd_b}
    \end{subfigure}
    \caption{Employing \emph{MMD} or \emph{L2} as the constraints of fully-connected features in the two-task scenario starting from ImageNet. (a): The performance drop or gain compared to \textit{Joint Training} on the old tasks; (b): The performance drop or gain compared to \textit{Finetuning} on the new tasks.}
    \label{fig:fc_l2mmd}
\end{figure}

\subsubsection{Constraints of fully connected features}
\label{sec:c_fc}

MMD measures the distance between data distributions after mapping them to the RKHS, which is useful for the high-level semantic feature alignment~\cite{long2015learning,long2017deep}.
Performance of the method that replaces the MMD constraints with the L2 norm is illustrated in Figure \ref{fig:fc_l2mmd}.
The results indicate that it is inferior to the proposed method in both performance preservation on the old tasks and accuracy on the new tasks.

We have also tried introducing a discriminator to play the adversarial minimax game with the fully connected features.
However, the discriminator easily distinguishes which task the features come from after only several epochs, so the adversarial game cannot continue.
The reason is due to that the fully connected features are high-level semantic features containing rich task-specific information and cannot adapt immediately to confuse the discriminator.

\subsubsection{Complexity and Effectiveness}
\label{sec:hyper_param}

Although our method introduces a trainable discriminator and additional hyperparameters, there is no need to worry about the extra complexity.
The discriminator here can be trained in an end-to-end way along with the backbone network, without extra tricks.
The discriminator is initialized between tasks, avoiding the storage cost for auto-encoders in \textit{EBLL} and importance weights in \textit{EWC}, \textit{SI} and \textit{MAS}.
Besides, we find our method is not very sensitive to hyperparameters.
For Eq.(\ref{eq:ours}), the selection of $\lambda_1$ is discussed in Section~\ref{sec:dist} while $\lambda_3$ is set to 1.0 in all of the above experiments.
We choose 1.0 for $\lambda_2$ in the two-task scenario while a smaller one in a longer sequence.
Network design and training strategies are available in Section~\ref{sec:imple}.

The key idea of our method lies in that the intermediate features of the old model are treated as multilevel soft targets and guide the training process in multiple stages.
Aligning the features generated by \emph{different networks} from the \emph{same data} will not prevent the network from learning specialized features.
Instead, by distilling the knowledge of previous networks to the new one and acting as regularizers, our method reaches higher accuracy on the new tasks even than \textit{Joint Training} and \textit{Finetuning}.
The absolute improvement is not very significant because there are upper bounds in lifelong learning scenarios, e.g., the accuracy of \textit{Joint Training} in the five-task scenario.
In fact, both the improvements in performance preservation on the old tasks and outstanding accuracy on the new tasks are impressive.

\section{Conclusion and Future Work}

Lifelong learning is still an open research problem.
In this paper, we focus on the incremental task scenario and propose an improved activation regularization lifelong learning method with adversarial feature alignment.
Both the low-level visual features and high-level semantic features serve as soft targets when training on new data.
Aligning these features generated by different models but from same data provides sufficient supervised information for the old tasks and help to reduce forgetting.
Additionally, the proposed method gains even better performance than finetuning on the new tasks due to the knowledge distillation and regularization phenomenon.
Extensive experiments in the incremental task scenarios are conducted, and the results show that our method outperforms previous ones in both accuracies on new tasks and performance preservation on old tasks.

There are several directions for further improvements.
For example, the commonly used target datasets in lifelong learning scenarios are usually small compared to ImageNet.
We would like to conduct more experiments using other kinds of backbone models with larger datasets, to evaluate the generalization ability of our methods.

\section*{Acknowledgments}
This work is supported by the National Key R\&D Program of China (2018YFB1003703), the National Natural Science Foundation of China (61521002), as well as the Beijing Key Lab of Networked Multimedia (Z161100005016051).

\bibliographystyle{ieee}
\bibliography{ref}

\end{document}